\pdfoutput=1
\documentclass[10pt,twocolumn,letterpaper]{article}

\usepackage{cvpr}
\usepackage{times}
\usepackage{epsfig}
\usepackage{graphicx}
\usepackage{amsmath}
\usepackage{amssymb}
\usepackage{float}
\newtheorem{lemma}{Lemma}
\newtheorem{definition}{Definition}
\usepackage{multirow,bigdelim}
\usepackage[toc,page]{appendix}

\usepackage[pagebackref=true,breaklinks=true,letterpaper=true,colorlinks,bookmarks=false]{hyperref}

\cvprfinalcopy 

\def\cvprPaperID{3164} 

\begin{document}
	
	\title{PointNetVLAD: Deep Point Cloud Based Retrieval for Large-Scale Place Recognition}
	
	\author{Mikaela Angelina Uy
		\quad
		Gim Hee Lee\\
		Department of Computer Science, National University of Singapore\\
		{\tt\small \{mikacuy,gimhee.lee\}@comp.nus.edu.sg}
	}
	
	\maketitle
	
	\begin{abstract}
		\vspace{-0.3cm}
		Unlike its image based counterpart, point cloud based retrieval for place recognition has remained as an unexplored and unsolved problem. This is largely due to the difficulty in extracting local feature descriptors from a point cloud that can  subsequently be encoded into a global descriptor for the retrieval task. In this paper, we propose the PointNetVLAD where we leverage on the recent success of deep networks to solve point cloud based retrieval for place recognition. Specifically, our PointNetVLAD is a combination/modification of the existing PointNet and NetVLAD, which allows end-to-end training and inference to extract the global descriptor from a given 3D point cloud.
		Furthermore, we propose the ``lazy triplet and quadruplet" loss functions that can achieve more discriminative and generalizable global descriptors to tackle the retrieval task.
		We create benchmark datasets for point cloud based retrieval for place recognition, and the experimental results on these datasets show the feasibility of our PointNetVLAD. 
		Our code and datasets are publicly available on the project website \footnote{\url{https://github.com/mikacuy/pointnetvlad.git}}.
	\end{abstract}
	
	\vspace{-0.5cm}
\section{Introduction}
\vspace{-0.1cm}
Localization addresses the question of ``where am I in a given reference map", and it is of paramount importance for robots such as self-driving cars \cite{Christian2017} and drones \cite{Fraundorfer2012} to achieve full autonomy. 
A common method for the localization problem is to first store a map of the environment as a database of 3D point cloud built from a collection of images with Structure-from-Motion (SfM) \cite{Hartley2004}, or LiDAR scans with Simultaneous Localization and Mapping (SLAM) \cite{Thrun:2005}. Given a query image or LiDAR scan of a local scene, we then search through the database to retrieve the best match that will tell us the exact pose of the query image/scan with respect to the reference map. 

A two-step approach is commonly used in image based localization 
\cite{Sattler16CVPR,Sattler15ICCV,Sattler-2017,Zeisl15ICCV}
- (1) place recognition \cite{cummins:2010,Cummins:2008:FPL:1377516.1377517,Milford12ICRA,Akihiko:2015,GalvezTRO12}, followed by (2) pose estimation \cite{haralick1991pose}. In place recognition, a global descriptor is computed for each of the images used in SfM by aggregating local image descriptors, e.g. SIFT, using the bag-of-words approach \cite{Nister:2006,SivicZ03}.
Each global descriptor is stored in the database together with the camera pose of its associated image with respect to the 3D point cloud reference map. Similar global descriptor is extracted from the query image and the closest global descriptor in the database can be retrieved via an efficient search. The camera pose of the closest global descriptor would give us a coarse localization of the query image with respect to the reference map. 
In pose estimation, we compute the exact pose of the query image with the Perspective-n-Point (PnP) \cite{haralick1991pose} and geometric verification \cite{LeeP14} algorithms.

\begin{figure}[t]
	\begin{center}
		\includegraphics[width=1\linewidth]{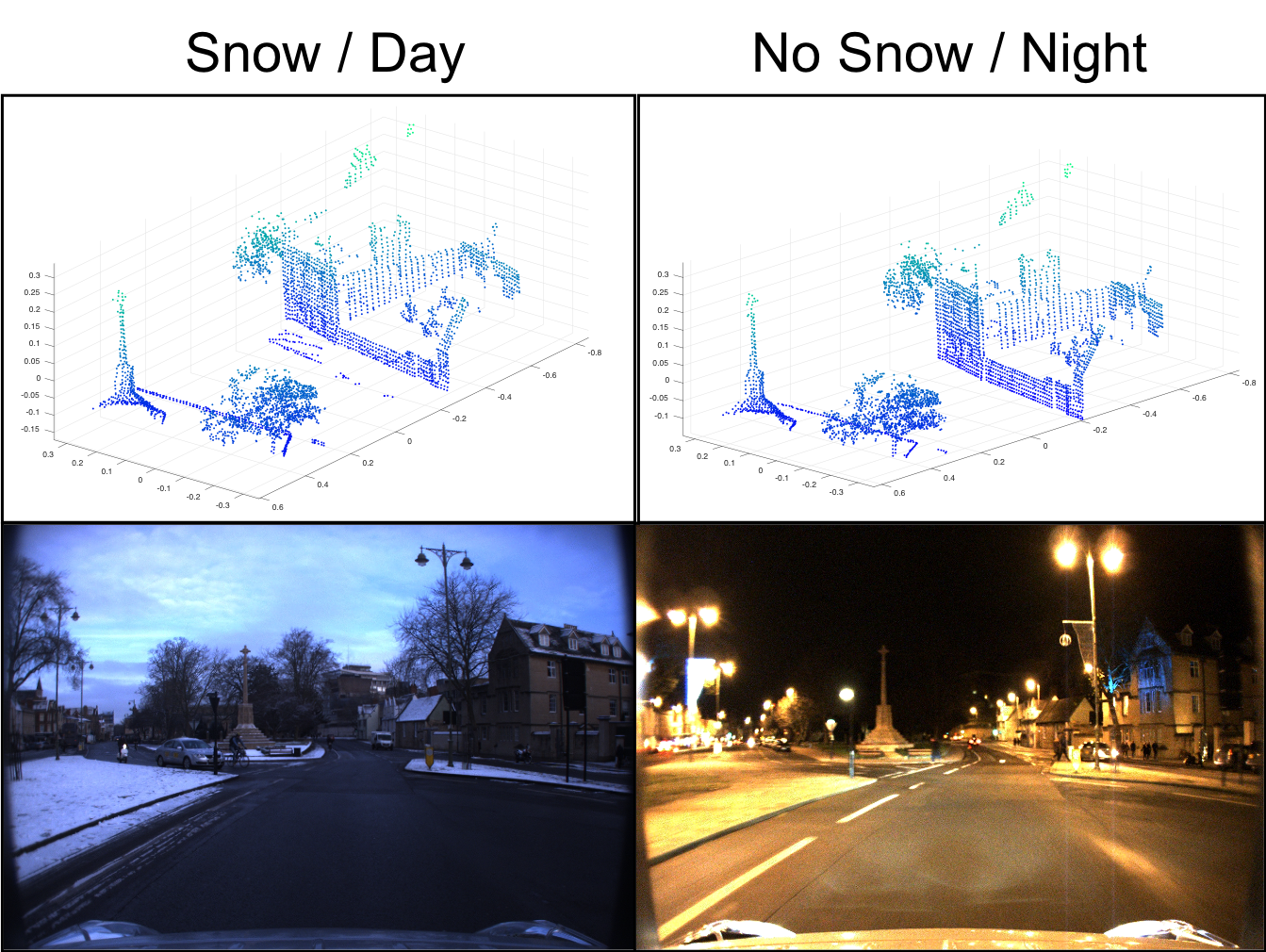}
	\end{center}
	\vspace{-0.3cm}
	\caption{
		Two pairs of 3D LiDAR point clouds (top row) and images (bottom row) taken from two different times. It can be seen that the pair of 3D LiDAR point cloud remain largely invariant to the lighting and seasonal changes that made it difficult to match the pair of images. Data from \cite{RobotCarDatasetIJRR}.\vspace{-0.5cm}}
	\label{fig:CmpLidarImg}
\end{figure}

The success of image based place recognition is largely attributed to the ability to extract image feature descriptors e.g. SIFT, that are subsequently aggregated with bag-of-words to get the global descriptor.
Unfortunately, there is no algorithm to extract local features similar to SIFT for LiDAR scans.
Hence, it becomes impossible to compute global descriptors from the bag-of-word approach to do LiDAR based place recognition. Most existing approaches circumvent this problem by using readings from the Global Positioning System (GPS) to provide coarse localization, followed by point cloud registration, e.g. the iterative closest point (ICP) \cite{Segal-RSS-09} or autoencoder based registration \cite{Elbaz_2017_CVPR}, for pose-estimation. As a result, LiDAR based localization is largely neglected since GPS might not be always available, despite the fact that much more accurate localization results can be obtained from LiDAR compared to images due to the availability of precise depth information. Furthermore, in comparison to images, the geometric information from LiDARs are invariant to drastic lighting changes, thus making it more robust to perform localization on queries and databases taken from different times of the day, e.g. day and night, and/or different seasons of the year. Fig. \ref{fig:CmpLidarImg} shows an example of a pair of 3D LiDAR point clouds and images that are taken from the same scene over two different times (daytime in winter on the left column, and nighttime in fall on the right column). It is obvious that the lighting (day and night) and seasonal (with and without snow) changes made it difficult even for human eye to tell that the pair of images (bottom row) are from the same scene. In contrast, the geometric structures of the LiDAR point cloud remain largely unchanged.  

In view of the potential that LiDAR point clouds could be better in the localization task, we propose the PointNetVLAD - a deep network for large-scale 3D point cloud retrieval to fill in the gap of place recognition in the 3D point cloud based localization. Specifically, our PointNetVLAD
is a combination of the existing PointNet \cite{qi2016pointnet} and NetVLAD \cite{Arandjelovic16}, which allows end-to-end training and inference to extract the global descriptor from a given 3D point cloud. We provide the proof that NetVLAD is a symmetric function, which is essential for our PointNetVLAD to achieve permutation invariance on the 3D point cloud input. We apply metric learning \cite{Chopra:2005} to train our PointNetVLAD to effectively learn a mapping function that maps input 3D point clouds to discriminative global descriptors. Additionally, we propose the ``lazy triplet and quadruplet" loss functions that achieve more generalizable global descriptors by maximizing the differences between all training examples from their respective hardest negative. We create benchmark datasets for point cloud based retrieval for place recognition based on the open-source Oxford RobotCar dataset \cite{RobotCarDatasetIJRR} and three additional datasets collected from three different areas with a Velodyne-64 LiDAR mounted on a car. Experimental results on the benchmark datasets verify the feasibility of our PointNetVLAD. 
	\section{Related Work}

%

Unlike the maturity of handcrafted local feature extraction for 2D images \cite{Lowe:2004,Bay:2008}, no similar methods proposed for 3D point cloud have reached the same level of maturity. 
In NARF~\cite{Steder:2010}, Steder $et.~al.$ proposed an interest point extraction algorithm for object recognition. In SHOT~\cite{Tombari:2010}, Tombari $et.~al.$  suggested a method to extract 3D descriptors for surface matching. However, both \cite{Steder:2010,Tombari:2010} rely on stable surfaces for descriptor calculation and are more suitable for dense rigid objects from 3D range images but not for outdoor LiDAR scans. A point-wise histogram based descriptor - FPFH was proposed in \cite{Rusu:2009,Radu:2008} for registration. It works on outdoor 3D data but requires high data density, thus making it not scalable to large-scale environments.
%

In the recent years, handcrafted features have been increasingly replaced by deep networks that have shown amazing performances. The success of deep learning has been particularly noticeable on 2D images where convolution kernels can be easily applied to the regular 2D lattice grid structure of the image. However, it is more challenging for convolution kernels to work on 3D points that are orderless.
Several deep networks attempt to mitigate this challenge by transforming point cloud inputs into regular 3D volumetric representations. Some of these works include:
3D ShapeNets \cite{Zhirong15CVPR} for recognition, volumetric CNNs \cite{Qi:2016} and OctNet \cite{Riegler2017OctNet} for classification. Additionally, 3DMatch \cite{Zeng:2017} learns local descriptors for small-scale indoor scenes and Vote3D \cite{Wang-RSS-15} is used for object detection on the outdoor KITTI dataset.
Instead of volumetric representation, MVCNN \cite{SuMKL:2015} projects the 3D point cloud into 2D image planes across multiple views to solve the shape recognition problem. 
Unfortunately, volumetric representations and 2D projections based deep networks that work well on object and small-scale indoor levels 
do not scale well for our large-scale outdoor place recognition problem.
%

It is not until the recent PointNet \cite{qi2016pointnet} that 
made it possible for direct input of 3D point cloud. 
The key to its success is the symmetric max pooling function that enables the aggregation of local point features into a latent representation which is invariant to the permutation of the input points. PointNet focuses on the classification task: shape classification and per-point classification (i.e. part segmentation, scene semantic parsing) on rigid objects and enclosed indoor scenes. PointNet is however not shown to do large-scale point cloud based place recognition. 
Kd-network \cite{Klokov:2017} also works for unordered point cloud inputs by transforming them into kd-trees. However, it is non-invariant/partially-invariant to rotation/noise that are both present in large-scale outdoor LiDAR point clouds. 
\begin{figure*}[t]
	\begin{center}
		\includegraphics[width=1\linewidth]{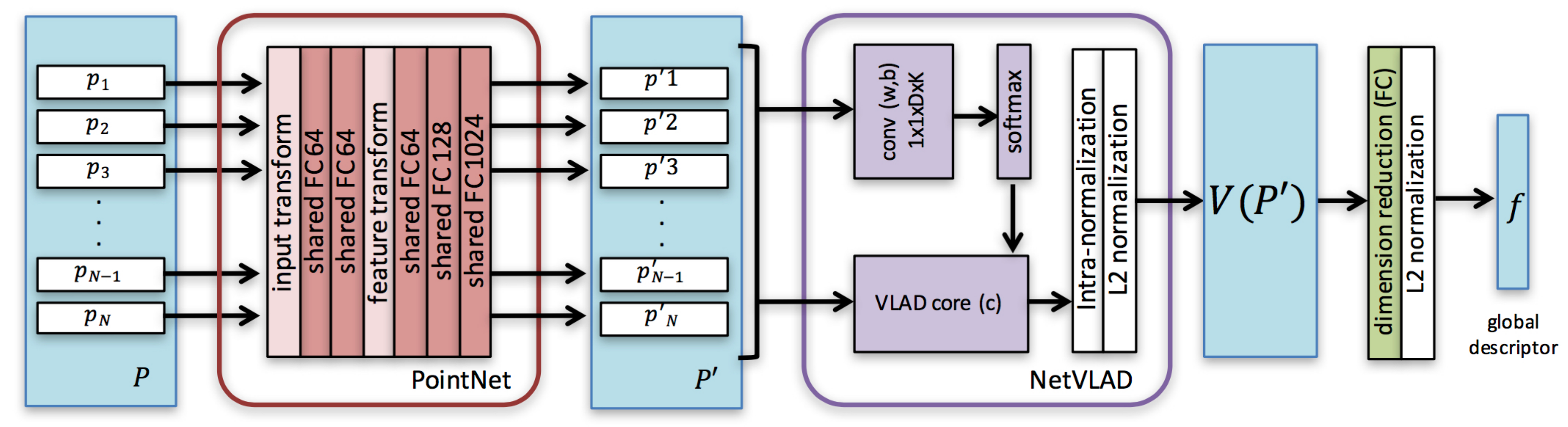}
	\end{center}
	\vspace{-0.5cm}
	\caption{Network architecture of our PointNetVLAD.\vspace{-0.4cm}}
	\label{fig:PointNetVLAD}
\end{figure*}

In \cite{Arandjelovic16}, Arandjelovi\'c $et.~al.$ proposed the NetVLAD - a deep network that models after the successful bag-of-words approach VLAD \cite{conf/cvpr/JegouDSP10, Arandjelovic:2013}. The NetVLAD is an end-to-end deep network made up of the VGG/Alexnet \cite{Simonyan:2016, Alex:2012} for local feature extraction, followed by the NetVLAD aggregation layer for clustering the local features into VLAD global descriptor. NetVLAD is trained on images obtained from the Google Street View Time Machine, a database consisting of multiple instances of places taken at different times, to perform the image based place recognition tasks. Results in \cite{Arandjelovic16} show that using the NetVLAD layer significantly outperformed the original non-deep learning based VLAD and its deep learning based max pooling counterpart. Despite the success of NetVLAD for image retrieval, it does not work for our task of point cloud based retrieval since it is not designed to take 3D points as input. 

Our PointNetVLAD leverages on the success of PointNet \cite{qi2016pointnet} and NetVLAD \cite{Arandjelovic16} to do 3D point cloud based retrieval for large-scale place recognition. Specifically, we show that our PointNetVLAD, which is a combination/modification of the PointNet and NetVLAD, originally used for point based classification and image retrieval respectively, is capable of doing end-to-end 3D point cloud based place recognition. 

	\section{Problem Definition}

Let us denote the reference map $\mathcal{M}$ as a database of 3D points defined with respect to a fixed reference frame. 
We further define that the reference map $\mathcal{M}$ is divided into a collection of $M$ submaps $\left\lbrace \mathsf{m}_1, ..., \mathsf{m}_M  \right\rbrace$ such that $\mathcal{M} = \bigcup_{i=1}^{M} \mathsf{m}_i$. The area of coverage ($AOC$) of all submaps are made to be approximately the same, i.e. $AOC(\mathsf{m}_1) \approx ...~AOC(\mathsf{m}_M)$, and the number of points in each submap is kept small, i.e. $|\mathsf{m}_i| \ll |\mathcal{M}|$. 
We apply a downsampling filter $\mathcal{G}(.)$ to ensure that the number of points of all downsampled submaps are the same, i.e. $|\mathcal{G}(\mathsf{m}_1)| = ...~|\mathcal{G}(\mathsf{m}_M)|$.
The problem of large-scale 3D point cloud based retrieval can be formally defined as follows:

\begin{definition}
Given a query 3D point cloud denoted as $\mathsf{q}$, where  $AOC(\mathsf{q}) \approx AOC(\mathsf{m}_i)$ and $|\mathcal{G}(\mathsf{q})| = |\mathcal{G}(\mathsf{m}_i)|$, our goal is to retrieve
the submap $\mathsf{m}_{*}$ from the database $\mathcal{M}$ that is structurally most similar to $\mathsf{q}$. 

\end{definition}

Towards this goal, we design a deep network to learn a function $f(.)$ that maps a given downsampled 3D point cloud $\bar{\mathsf{p}}={\mathcal{G}(\mathsf{p})}$, where $AOC(\mathsf{p}) \approx AOC(\mathsf{m}_i)$, to a fixed size global descriptor vector $f(\bar{\mathsf{p}})$ such that $d(f(\bar{\mathsf{p}}),f(\bar{\mathsf{p}}_r)) < d(f(\bar{\mathsf{p}}),f(\bar{\mathsf{p}}_s))$, if $\mathsf{p}$ is structurally similar to $\mathsf{p}_r$ but dissimilar to $\mathsf{p}_s$. $d(.)$ is some distance function, e.g. Euclidean distance function. Our problem then simplifies to finding the submap $\mathsf{m}_* \in \mathcal{M}$ such that its global descriptor vector $f(\bar{\mathsf{m}}_*)$ gives the minimum distance with the global descriptor vector $f(\bar{\mathsf{q}})$ from the query $\mathsf{q}$, i.e. $d(f(\bar{\mathsf{q}}),f(\bar{\mathsf{m}}_*)) < d(f(\bar{\mathsf{q}}),f(\bar{\mathsf{m_i}})), \forall i \neq *$. In practice, this can be done efficiently by a simple nearest neighbor search through a list of global descriptors $\{f(\bar{\mathsf{m}}_i) \mid i\in1,2,..,M\}$ that can be computed once offline and stored in memory, while  $f(\bar{\mathsf{q}})$ is computed online. 


	\section{Our PointNetVLAD}

In this section, we will describe the network architecture of PointNetVLAD and the loss functions that we designed to learn the function 
$f(.)$ that maps a downsampled 3D point cloud to a global descriptor. We also show the proof that the NetVLAD layer is permutation invariant, thus suitable for 3D point cloud. 
 
\subsection{The Network Architecture}

Fig.~\ref{fig:PointNetVLAD} shows the network architecture of our PointNetVLAD, which is made up of three main components - (1) PointNet \cite{qi2016pointnet}, (2) NetVLAD \cite{Arandjelovic16} and (3) a fully connected network. Specifically, we take the first part of PointNet, cropped just before the maxpool aggregation layer. The input to our network is the same as PointNet, which is a point cloud made up of a set of 3D points, $P = \left\lbrace p_1, ..., p_N \mid p_n \in \mathbb{R}^3 \right\rbrace$. Here, we denote $P$ as a fixed size point cloud after applying the filter $\mathcal{G}(.)$; we drop the bar notation on $P$ for brevity.
The role of PointNet is to map each point in the input point cloud into a higher dimensional space, i.e. $P = \left\lbrace p_1, ..., p_N \mid p_n \in \mathbb{R}^3 \right\rbrace \longmapsto P' = \left\lbrace p'_1, ..., p'_N \mid p'_n \in \mathbb{R}^D\right\rbrace$, where $D \gg 3$. Here, PointNet can be seen as the component that learns to extract a $D$-dimensional local feature descriptor from each of the input 3D points.

We feed the output local feature descriptors from PointNet as input to the NetVLAD layer. The NetVLAD layer is originally designed to aggregate local image features learned from VGG/AlexNet into the VLAD bag-of-words global descriptor vector. By feeding the local feature descriptors of a point cloud into the layer, we create a machinery that generates the global descriptor vector for an input point cloud. The NetVLAD layer learns $K$ cluster centers, i.e. the visual words, denoted as $\{c_1, ..., c_K \mid c_k\in\mathbb{R}^{D}\}$, and outputs a ($D \times K$)-dimensional vector $V(P')$. The output vector $V(P') = [V_1(P'), ..., V_K(P')]$ is an aggregated representation of the local feature vectors, where $V_k(P') \in \mathbb{R}^D$ is given by:
\begin{equation}\label{eq:NetVLAD}
V_{k}(P')=\sum_{i=1}^{n}{\frac{e^{w_k^{T}p'_i+b_k}}{\sum_{k'}e^{w_{k'}^{T}p'_i+b_{k'}}}(p_i'-c_k)}.
\end{equation}
\noindent $\{w_k\}$ and $\{b_k\}$ are the weights and biases that determine the contribution of local feature vector $p'_i$ to $V_k(p')$. All the weight and bias terms are learned during training.

The output from the NetVLAD layer is the VLAD descriptor \cite{conf/cvpr/JegouDSP10,Arandjelovic:2013} for the input point cloud. However, the VLAD descriptor is a high dimensional vector, i.e. ($D \times K$)-dimensional vector, that makes it computationally expensive for nearest neighbor search. To alleviate this problem, 
we use a fully connected layer to compress the $(D \times K)$ vector into a compact output feature vector, which is then L2-normalized to produce the final global descriptor vector $f(P) \in \mathbb{R}^{\mathcal{O}}$, where $\mathcal{O} \ll (D \times K)$,  for point cloud $P$ that can be used for efficient retrieval.


\subsection{Metric Learning}

We train our PointNetVLAD end-to-end to learn the function $f(.)$
that maps an input point cloud $P$ to a discriminative compact global descriptor vector $f(P)\in\mathbb{R}^{\mathcal{O}}$, where $\|f(P)\|_{2}=1$. To this end, we 
propose the ``Lazy Triplet" and ``Lazy Quadruplet" losses that can learn discriminative and generalizable global descriptors. We obtain a set of training tuples from the training dataset, where each tuple is denoted as $\mathcal{T} = (P_a,P_{pos},\{P_{neg}\})$. $P_a$, $P_{pos}$ and $\{P_{neg}\}$ denote an anchor point cloud, a structurally similar (``positive'') point cloud to the anchor and a set of structurally dissimilar (``negative'') point clouds to the anchor, respectively. 
The loss functions are designed to 
minimize the distance between the global descriptor vectors of $P_a$ and $P_{pos}$, i.e. $\delta_{pos}=d(f(P_a),f(P_{pos}))$, and maximize the distance between the global descriptor vectors of $P_a$ and some $P_{neg_{j}}\in \{P_{neg}\}$, i.e. $\delta_{neg_{j}}=d(f(P_a),f(P_{neg_{j}}))$. $d(.)$ is a predefined distance function, which we take to be the squared Euclidean distance in this work.
\newline

\noindent \textbf{Lazy triplet:} 
For each training tuple $\mathcal{T}$, our lazy triplet loss focuses on maximizing the distance between $f(P_a)$ and the global descriptor vector of the
closest/hardest negative in $\{P_{neg}\}$, denoted as $f(P_{neg_j}^{-})$. Formally, the lazy triplet loss is defined as
\begin{equation}
\mathcal{L}_{lazyTrip}(\mathcal{T})=\max_{j}([\alpha+\delta_{pos}-\delta_{neg_{j}}]_{+}),
\label{eq:lazyTriplet}
\end{equation}
where $[\ldots]_{+}$ denotes the hinge loss and $\alpha$ is a constant parameter giving the margin.   
The max operator selects the closest/hardest negative $P_{neg_j}^{-}$ in $\{P_{neg}\}$ that gives the smallest $\delta_{neg_j}$ value in a particular iteration. Note that $P_{neg_j}^{-}$ of each training tuple changes because the parameters of the network that determine $f(.)$ get updated during training, hence a different point cloud in $\{P_{neg}\}$ might get mapped to a global descriptor that is nearest to $f(P_a)$ at each iteration. Our choice to iteratively 
use the closest/hardest negatives over all training tuples ensures that the network learns from all the hardest examples to get a more discriminative and generalizable function $f(.)$.
\newline
 
\begin{figure*}[h]
	\begin{center}
		\includegraphics[width=1\linewidth]{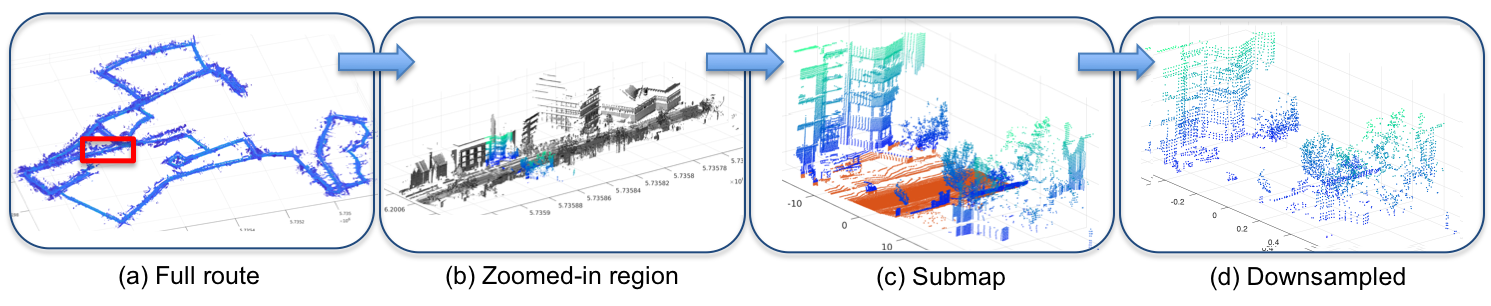}
	\end{center}
	\vspace{-0.3cm}
	\caption{Dataset preprocessing: (a) A full route from the Oxford RobotCar dataset. (b) Zoomed-in region of the 3D point cloud in the red box shown in (a). (c) An example of submap with the detected ground plane shown as red points. (d) A downsampled submap that is centered at origin and all points within [-1,1]m.\vspace{-0.4cm}}
	\label{fig:dataProcessing}
\end{figure*}

\noindent \textbf{Lazy quadruplet:} The choice to maximize the distance between $f(P_a)$ and $f(P_{neg_j}^{-})$ might lead to an undesired reduction of the distance between $f(P_{neg_j}^{-})$ and another point cloud $f(P_{false})$, where $P_{false}$ is structurally dissimilar to $P_{neg_j}^{-}$.
To alleviate this problem, we maximize an additional distance $\delta_{neg_k^*}=d(f(P_{neg^*}),f(P_{neg_k}))$, where $P_{neg^*}$ is randomly sampled from the training dataset at each iteration and is dissimilar to all point clouds in $\mathcal{T}$.
The lazy quadruplet loss is defined as
\begin{equation}
\begin{split}
\mathcal{L}_{lazyQuad}(\mathcal{T},P_{neg^*})=&\max_{j}([\alpha+\delta_{pos}-\delta_{neg_{j}}]_{+})\\
+&\max_{k}([\beta+\delta_{pos}-\delta_{neg_{k}^*}]_{+}),
\end{split}
\label{fig:lazyQuad}
\end{equation}

\noindent where $\beta$ is a another constant parameter giving the margin. The max operator of the second term selects the hardest negative $P_{neg_k}^{-}$ in $\{P_{neg}\}$ that give the smallest $\delta_{neg_k}$ value.
\newline

\noindent \textbf{Discussion:} Original triplet and quadruplet losses use the sum instead of the max operator proposed in our ``lazy" variants. These losses have been shown to work well for different applications such as facial recognition \cite{DBLP:journals/corr/SchroffKP15, DBLP:journals/corr/ChenCZH17}. However, maximizing $\delta_{neg_{j}}$ for all $\{P_{neg}\}$ leads to a compounding effect where the contribution of each negative training data diminishes
as compared to the contribution from a single hardest negative training data. As a result, the original triplet and quadruplet losses tend to take longer to train, and lead to a less discriminative function $f(.)$ that produces inaccurate retrieval results.
Experimental results indeed show that both our ``lazy" variants outperform the original losses by a competitive margin with the lazy quadruplet loss slightly outperforming the lazy triplet loss.

\subsection{Permutation Invariance}
Unlike its image counterpart, a set of points in a point cloud are unordered. Consequently, a naive design of the network could produce different results from different orderings of the input points.  
It is therefore necessary for the network to be input order invariant for it to be suitable for point clouds. This means that the network will output the same global descriptor $f(P)$ for point cloud $P$ regardless of the order in which the points in $P$ are arranged. We rigorously show that this property holds for PointNetVLAD.

Given an input point cloud $P=\{p_1,p_2,\ldots,p_N\}$, the layers prior to NetVLAD, i.e. PointNet, transform each point in $P$ independently into $P'=\{p'_1,p'_2,\ldots,p'_N\}$, hence it remains to show that NetVLAD is a symmetric function, which means that its output $V(P')$ would be invariant to the order of the points in $P'$ leading to an output global descriptor $f(P')$ that is order invariant.
\begin{lemma}
	NetVLAD is a symmetric function
\end{lemma}
\textbf{Proof:} Given the feature representation of input point cloud $P$ as $\{p'_1,p'_2,\ldots,p'_N\}$, we have the output vector $V=[V_1,V_2,\ldots,V_K]$ of NetVLAD such that $\forall k$,
\begin{equation} \label{eq2}
V_k = h_k(p'_1)+h_k(p'_2)+\ldots+h_k(p'_N)=\sum_{t=1}^{N}h_k(p'_{t}),
\end{equation}
where
\begin{equation}\label{eq3}
h_k(p')=\frac{e^{w_k^{T}p'+b_k}}{\sum_{k'}e^{w_{k'}^{T}p'+b_{k'}}}(p'-c_k).
\end{equation}
Suppose we have another point cloud $\tilde{P}=\{p_1,\ldots,p_{i-1},p_j,p_{i+1},\ldots,p_{j-1},p_{i},p_{j+1},\ldots,p_N\}$~that is similar to $P$ except for reordered points $p_i$~and~$p_j$. Then the feature representation of $\tilde{P}$ is given by $\{p'_1,\ldots,p'_{i-1},p'_j,p'_{i+1},\ldots,p'_{j-1},p'_{i},p'_{j+1},\ldots,p'_N\}$. Hence $\forall k$, we have
\begin{equation}
\begin{split}
\tilde{V}_k=&h_k(p'_1)+\ldots+h_k(p'_{i-1})+\\
&h_k(p'_{j})+h_k(p'_{i+1})+\ldots+h_k(p'_{j-1})+\\
&h_k(p'_{i})+h_k(p'_{j+1})+\ldots+h_k(p'_N)\\
=&\sum_{t=1}^{N}h_k(p'_{t}) = V_k.
\end{split}
\end{equation}
Thus, $f(P)=f(\tilde{P})$ and completes our proof for symmetry.

	\section{Experiments}
\vspace{-0.1cm}
\subsection{Benchmark Datasets}
\vspace{-0.1cm}
We create four benchmark datasets suitable for LiDAR-based place recognition to train and evaluate our network: one from the open-source Oxford RobotCar \cite{RobotCarDatasetIJRR} and three in-house datasets of a university sector (U.S.), a residential area (R.A.) and a business district (B.D.).
These are created using a LiDAR sensor mounted on a car that repeatedly drives through each of the four regions at different times traversing a 10km, 10km, 8km and 5km route on each round of Oxford, U.S., R.A. and B.D., respectively. For each run of each region, the collected LiDAR scans are used to build a unique reference map of the region. The reference map is then used to construct a database of submaps that represent unique local areas of the region for each run. Each reference map is built with respect to the UTM coordinate frame using GPS/INS readings. 
\newline
\vspace{-0.3cm}

\noindent\textbf{Submap preprocessing} The ground planes are removed in all submaps since they are non-informative and repetitive structures. The resulting point cloud is then downsampled to 4096 points using a voxel grid filter \cite{Rusu_ICRA2011_PCL}. Next, it is shifted and rescaled to be zero mean and inside the range of [-1, 1]. Each downsampled submap is tagged with a UTM coordinate at its respective centroid, thus allowing supervised training and evaluation of our network. To generate training tuples, we define structurally similar point clouds to be at most 10m apart and those structurally dissimilar to be at least 50m apart. Fig.~\ref{fig:dataProcessing} shows an example of a reference map, submap and downsampled submap.
\newline
\vspace{-0.3cm}

\noindent\textbf{Data splitting and evaluation} 
We split each run of each region of the datasets into two disjoint reference maps used for training and testing. We further split each reference map into a set of submaps at regular intervals of the trajectory in the reference map. Refer to the supplementary material for more details on data splitting. We obtain a total of 28382 submaps for training and 7572 submaps for test from the Oxford and in-house datasets. To test the performance of our network, we use a submap from a testing reference map as a query point cloud and all submaps from another reference map of a different run that covers the same region as the database. The query submap is successfully localized if it retrieves a point cloud within 25m.
\newline
%
\vspace{-0.3cm}

\noindent\textbf{Oxford Dataset} We use 44 sets of full and partial runs from the Oxford RobotCar dataset \cite{RobotCarDatasetIJRR}, which were collected at different times with a SICK LMS-151 2D LiDAR scanner. Each run is geographically split into $70\%$ and $30\%$ for the training and testing reference maps, respectively. We further split each training and testing reference map into submaps at fixed regular intervals of 10m and 20m, respectively. Each submap includes all 3D points that are within a 20m trajectory of the car. This resulted in 21,711 training submaps, which are used to train our baseline network, and 3030 testing submaps ($\sim$ 120-150 submaps/run). 
\newline
\vspace{-0.3cm}

\noindent\textbf{In-house Datasets} The three in-house datasets are constructed from Velodyne-64 LiDAR scans of five different runs of each of the regions U.S., R.A. and B.D. that were collected at different times. These are all used as testing reference maps to test the generalization of our baseline network trained only on Oxford. Furthermore, we also geographically split each run of U.S. and R.A. into training and testing reference maps, which we use for network refinement. Submaps are taken at regular intervals of 12.5m and 25m for each training and testing reference maps, respectively. All 3D points within a 25m$\times$25m bounding box centered at each submap location are taken. Table \ref{tab:dataset} shows the breakdown on the number of training and testing submaps used in the baseline and refined networks.

\begin{table}[h]
	\centering\small
	\begin{tabular}{c|c|l|l|l}
		\multirow{2}{*}{}&\multicolumn{2}{c|}{Training$^+$}&\multicolumn{2}{c}{Test$^\times$}\\
		\cline{2-5}
		&Baseline&\multicolumn{1}{c|}{Refine}&\multicolumn{1}{c|}{Baseline}&\multicolumn{1}{c}{Refine}\\
		\hline
		Oxford&21711&\multicolumn{1}{c|}{21711}&\multicolumn{1}{c|}{3030}&\multicolumn{1}{c}{3030}\\
		U.S.&-&\multicolumn{1}{c|}{\rdelim\}{2}{10pt}\multirow{2}{*}{6671}}&400$^*$\rdelim\}{3}{10pt}\multirow{3}{*}{4542}&80$^*$ \rdelim\}{3}{10pt}\multirow{3}{*}{1766}\\
		R.A.&-&&320$^*$ &75$^*$ \\
		B.D.&-&-&200$^*$ &200$^*$ \\
	\end{tabular}
	\caption{Number of training and testing submaps for our baseline and refined networks. $^*$approximate number of submaps/run is given because the number of submaps differ slightly between each run; $^+$overlapping and $^\times$disjoint submaps.}
	\label{tab:dataset}
\end{table}
%
	\begin{figure}[t]
	\begin{center}
		\includegraphics[width=1\linewidth]{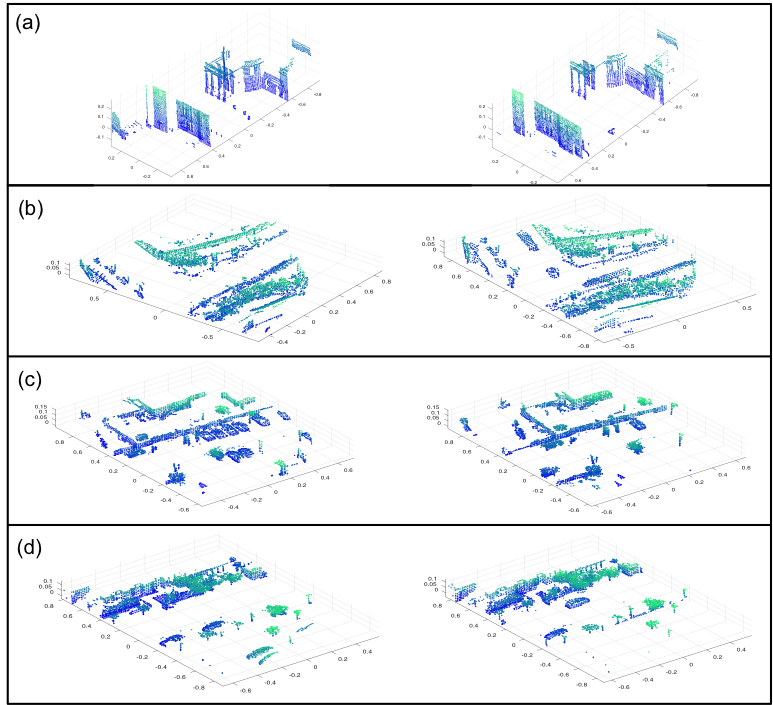}
	\end{center}
	\vspace{-0.3cm}
	\caption{Sample point clouds from (a) Oxford, (b) U.S., (c) R.A. and (d) B.D., respectively: left shows the query submap and right shows the successfully retrieved corresponding point cloud.}
	\label{fig:oxford_match}
\end{figure}
\vspace{-0.2cm}
\subsection{Results}
%
%
\begin{figure*}[t]
	\begin{center}
		\includegraphics[width=1\linewidth]{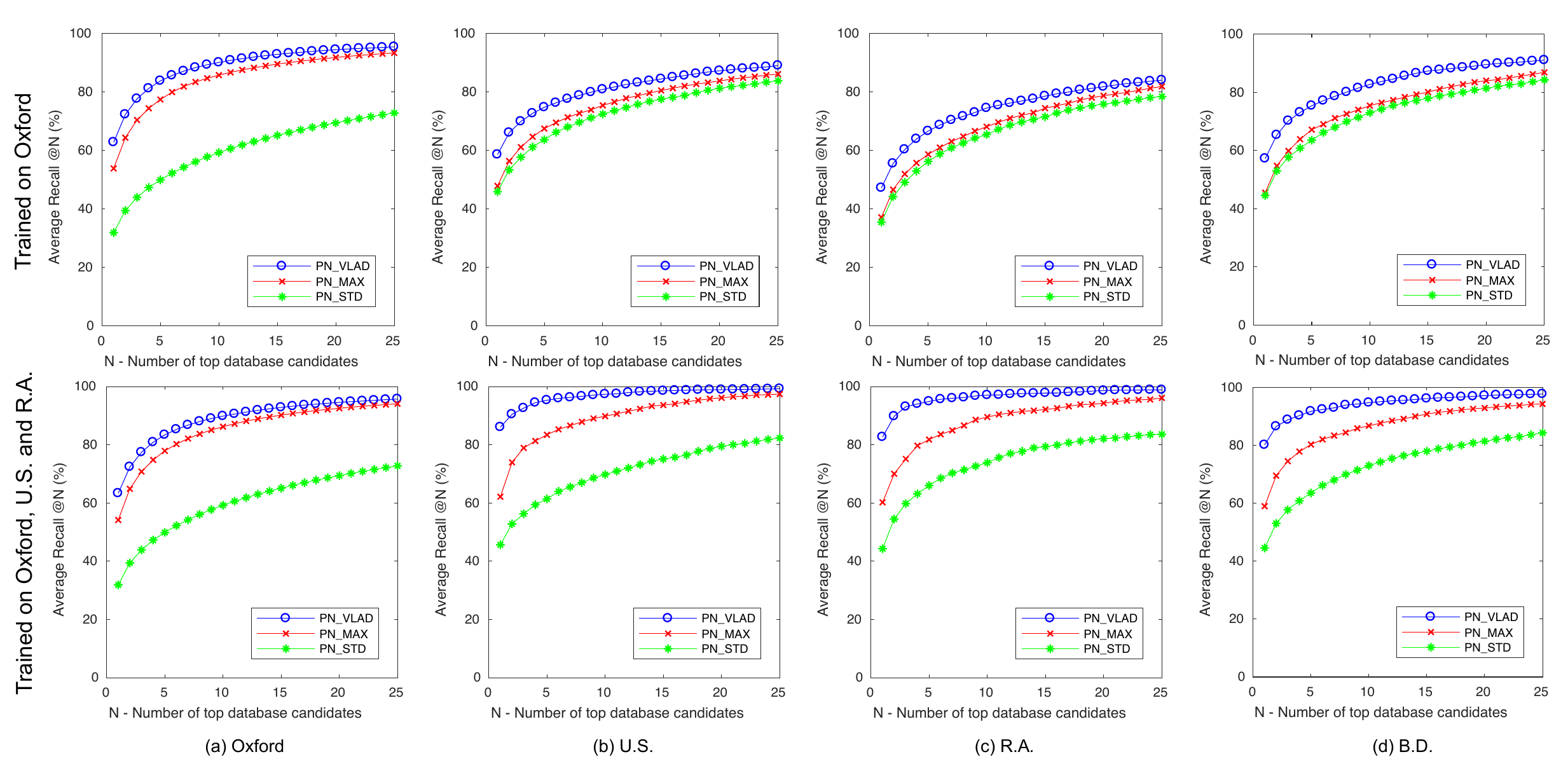}
	\end{center}
	\vspace{-0.4cm}
	\caption{\textbf{Average recall of the networks.} Top row shows the average recall when PN\_VLAD and PN\_MAX were only trained on Oxford. Bottom row shows the average recall when PN\_VLAD and PN\_MAX were trained on Oxford, U.S. and R.A. \vspace{-0.4cm}}
	\label{fig:results_recall}
\end{figure*}

We present results to show the feasibility of our PointNetVLAD (PN\_VLAD) for large-scale point cloud based place recognition. Additionally, we compare its performance to the original PointNet architecture with the maxpool layer (PN\_MAX) and a fully connected layer to produce a global descriptor with output dimension equal to ours; this is also trained end-to-end for the place recognition task. Moreover, we also compare our network with the state-of-the-art PointNet trained for object classification on rigid objects in ModelNet (PN\_STD) to investigate whether the model trained on ModelNet can be scaled to large-scale environments. We cut the trained network just before the softmax layer hence producing a 256-dim output vector. 
\newline
\begin{table}
	\begin{center}
		\begin{tabular}{|l|c|c|c|}
			\hline
			& PN\_VLAD & PN\_MAX & PN\_STD\\
			\hline
			Oxford& \textbf{80.31} & 73.44 & 46.52  \\
			\hline
			U.S.& \textbf{72.63} & 64.64 & 61.12 \\
			\hline
			R.A.& \textbf{60.27} & 51.92 & 49.07 \\
			\hline
			B.D.& \textbf{65.30} & 54.74 & 53.02\\
			\hline			
		\end{tabular}
	\end{center}
	\vspace{-0.2cm}
	\caption{Baseline results showing the average recall (\%) at top 1\% for each of the models.\vspace{-0.2cm}}
	\label{tab:baseline}
\end{table}

\vspace{-0.7cm}
\noindent\textbf{Baseline Networks} 
We train the PN\_STD, PN\_MAX and our PN\_VLAD using only the Oxford training dataset. The network configurations of PN\_STD and PN\_MAX are set to be the same as \cite{qi2016pointnet}. 
The dimension of the output global descriptor of PN\_MAX is set to be same as our PN\_VLAD, i.e. 256-dim.
Both PN\_MAX and our PN\_VLAD are trained with the lazy quadruplet loss, where we set the margins $\alpha=0.5$ and $\beta=0.2$. Furthermore, we set the number of clusters in our PN\_VLAD to be $K=64$. 
We test the trained networks on Oxford. The Oxford RobotCar dataset is a challenging dataset due to multiple roadworks that caused some scenes to change almost completely. We verify the generalization of our network by testing on completely unseen environments with our in-house datasets. Table \ref{tab:baseline} shows the top1\% recall of the different models on each of the datasets. 
It can be seen that PN\_STD does not generalize well for large scale place retrieval, and PN\_MAX does not generalize well to the new environments as compared to our PN\_VLAD. Fig. \ref{fig:results_recall} (top row) shows the recall curves of each model for the top 25 matches from each database pair for the four test datasets, where our network outperforms the rest. Note that the recall rate is the average recall rate of all query results from each submap in the test data. 
\newline
\begin{table}
	\begin{center}
		\begin{tabular}{|l|c|c|c|c|c|c|}
			\hline
			&\multicolumn{3}{c|}{PN\_VLAD}&\multicolumn{3}{c|}{PN\_MAX}\\
			\cline{2-7}
			&\small{D-128}&\small{D-256}&\small{D-512}&\small{D-128}&\small{D-256}&\small{D-512}\\
			\hline
			\small{Ox.}&\textbf{74.60}&80.31&80.33&71.93&73.44&\textbf{74.79}\\
			\hline
			\small{U.S.}&\textbf{66.03}&72.63&76.24&61.15&64.64&\textbf{65.79}\\
			\hline
			\small{R.A.}&\textbf{53.86}&60.27&63.31&49.25&51.92&\textbf{52.32}\\
			\hline
			\small{B.D.}&\textbf{59.84}&65.30&66.75&53.25&54.74&\textbf{56.63}\\
			\hline									
		\end{tabular}
	\end{center}
	\vspace{-0.1cm}
	\caption{Average recall (\%) at top1\% on the different datasets for output dimensionality analysis of PN\_VLAD and PN\_MAX. All models were trained on Oxford. Here, D- refers to global descriptors with output length D-dim.\vspace{-0.2cm}}
	\label{tab:dimension}
\end{table}

\begin{table}
	\begin{center}
		\begin{tabular}{|l|c|}
			\hline
			& Average recall\\
			\hline
			Triplet Loss& 71.20\\
			\hline
			Quadruplet Loss& 74.13\\
			\hline
			Lazy Triplet Loss& \textbf{78.99}\\
			\hline
			Lazy Quadruplet Loss& \textbf{80.31}\\
			\hline			
		\end{tabular}
	\end{center}
	\vspace{-0.1cm}
	\caption{Results representing the average recall (\%) at top1\% of PN\_VLAD tested and trained using different losses on Oxford.\vspace{-0.3cm} }
	\label{tab:losses}
\end{table}
\vspace{-0.2cm}
\noindent\textbf{Output dimensionality analysis} We study the discriminative ability of our network over different output dimensions of global descriptor $f$ for both our PN\_VLAD and PN\_MAX. As show in Table \ref{tab:dimension}, 
the performance of our PN\_VLAD with output length of 128-dim is on par with PN\_MAX with output length of 512-dim on Oxford, and marginally better on our in-house datasets.
The performance of our network increases from the output dimension of 128-dim to 256-dim, but did not increase further from 256-dim to 512-dim. Hence, we chose to use an output global descriptor of 256-dim in most of our experiments.
\newline
\vspace{-0.2cm}

\noindent\textbf{Comparison between losses} We compared our network's performance when trained on different losses. As shown in Table \ref{tab:losses}, our network performs better when trained on our lazy variants of the losses. Hence we chose to use the lazy quadruplet loss to train our PN\_VLAD and PN\_MAX. 
\newline
\begin{table}
	\begin{center}
		\begin{tabular}{|l|c|c|c|c|c|c|}
			\hline
			&\multicolumn{3}{c|}{Ave recall @1\%}&\multicolumn{3}{c|}{Ave recall@1}\\
			\cline{2-7}
			&\small{PN\_}&\small{PN\_}&\small{PN\_}&\small{PN\_}&\small{PN\_}&\small{PN\_}\\
			&\small{VLAD}&\small{MAX}&\small{STD}&\small{VLAD}&\small{MAX}&\small{STD}\\
			\hline
			\small{Ox.}&80.09&73.87&46.52&63.33&54.16&31.87\\
			\hline
			\small{U.S.}&90.10&79.31&56.95&86.07&62.16&45.67\\
			\hline
			\small{R.A.}&93.07&75.14&59.81&82.66&60.21&44.29\\
			\hline
			\small{B.D.}&\textbf{86.49}&69.49&53.02&\textbf{80.11}&58.95&44.54\\
			\hline									
		\end{tabular}
	\end{center}
	\caption{Final results showing the average recall (\%) at top 1\% (@1\%) and at top 1 (@1) after training on Oxford, U.S. and R.A.\vspace{-0.4cm}}
	\label{tab:final}
\end{table}
\vspace{-0.2cm}

\noindent\textbf{Network refinement} We further trained our network with U.S. and R.A. in addition to Oxford. 
This improves the generalizability of our network on the unseen data B.D. as can be seen from the last row of 
Table \ref{tab:final}  and second row of Fig. \ref{fig:results_recall}-(d).
We have shown the feasibility and potential of our PointNetVLAD for LiDAR based place recognition by achieving reasonable results despite the smaller database size compared to established databases for image based place recognition (\eg Google Street View Time Machine and Tokyo 24/7 \cite{Akihiko:2015}). We believe that given more publicly available LiDAR datasets suitable for place recognition our network can further improve its performance and bridge the gap of place recognition in LiDAR based localization.
\newline

\begin{figure}[t]
	\begin{center}
		\includegraphics[width=1\linewidth]{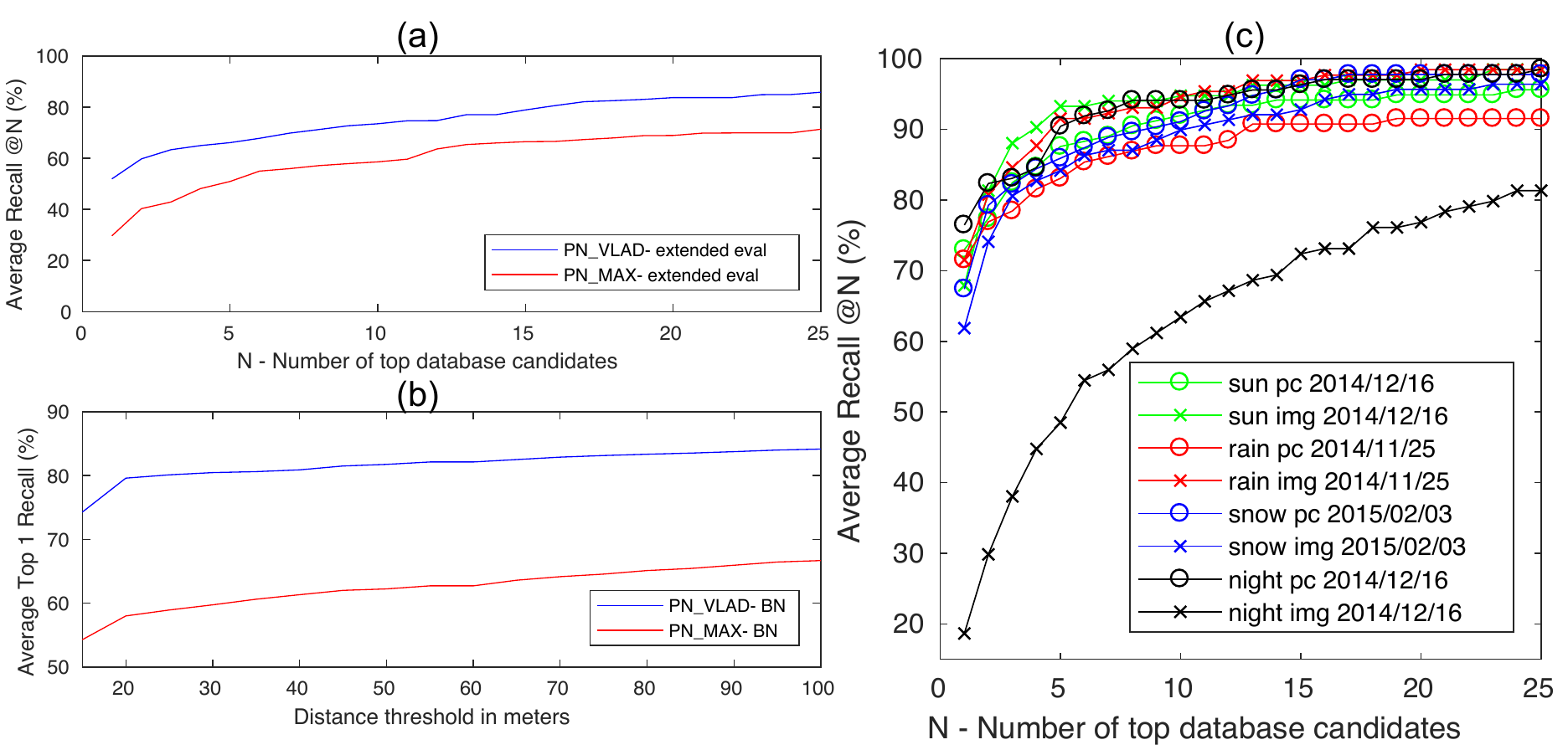}
	\end{center}
	\vspace{-0.4cm}
	\caption{(a) Average recall @N for retrieval from all reference areas. (b) Average recall at B.D. with varying distance thresholds. (c) Average recall @N with point clouds (pc) and images (img) as queries under various scene conditions, and retrieving from an overcast database in Oxford dataset.\vspace{-0.4cm}}
	\label{fig:add_results}
\end{figure}

\noindent\textbf{Extended evaluation}
Fig. \ref{fig:add_results}-(a) shows the average recall when queries from Oxford, U.S., R.A. and B.D. are retrieved from an extended database containing all four areas ($\sim33$km). Moreover, Fig. \ref{fig:add_results}-(b) shows the top 1 recall on unseen data B.D. with varying distance thresholds. It can be seen that on these extended evaluation metrics, our PN\_VLAD still outperforms PN\_MAX.
\newline

\vspace{-0.2cm}
\noindent\textbf{Image based comparisons under changing scene conditions}
We compare the performance of our point cloud based approach to the image based counterpart. We train NetVLAD according to the specifications specified in \cite{Arandjelovic16} with images from the center stereo camera of \cite{RobotCarDatasetIJRR}. These images are taken at the corresponding location of each point cloud submap used to train our PN\_VLAD. Fig. \ref{fig:add_results}-(c) shows retrieval results when query was taken from various scene conditions against an overcast database in the Oxford dataset. The performance of image based NetVLAD is comparable to our point cloud based PN\_VLAD in all cases, except for overcast (day) to night retrieval (a well-known difficult problem for image based methods) where our PN\_VLAD significantly outperforms NetVLAD. It can be seen that the use of point clouds makes the performance more robust to scene variations as they are more invariant to illumination and weather changes.
\newline

\begin{figure}[t]
	\begin{center}
		\includegraphics[width=1\linewidth]{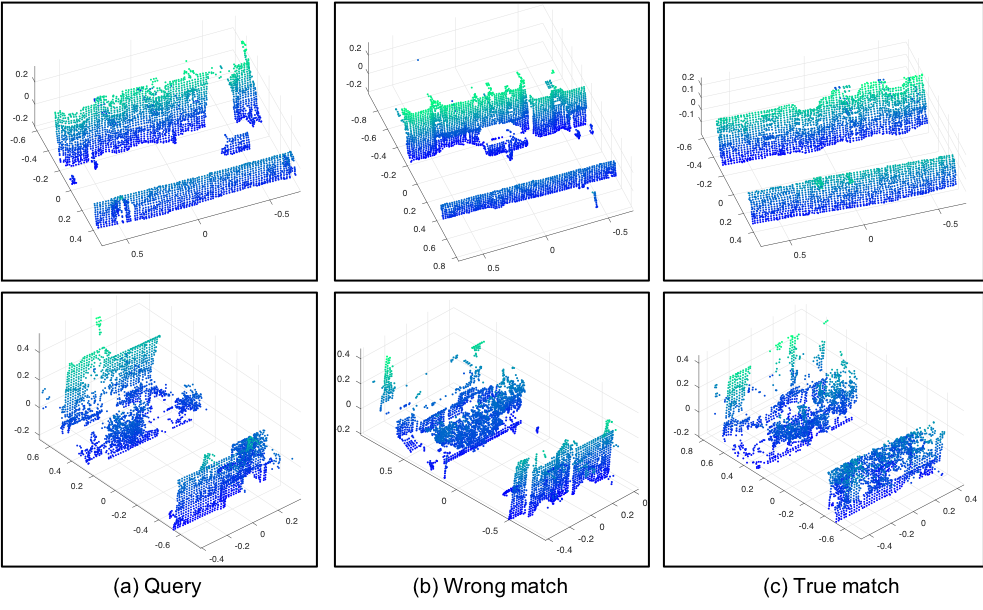}
	\end{center}
	\vspace{-0.4cm}
	\caption{Network limitations: These are examples of unsuccessfully retrieved point clouds by our network, where (a) shows the query, (b) shows the incorrect match to the query and (c) shows the true match.\vspace{-0.2cm}}
	\label{fig:mismatch}
\end{figure}

\vspace{-0.2cm}
\noindent\textbf{Qualitative Analysis} Fig. \ref{fig:CmpLidarImg} and \ref{fig:oxford_match} show some of the successfully recognized point clouds, and it can be seen that our network has learned to ignore irrelevant noise such as ground snow and cars (both parked and moving). Fig. \ref{fig:mismatch} shows examples of unsuccessfully retrieved point clouds, and we can see that our network struggles on continuous roads with very similar features (top row) and heavily occluded areas (bottom row). 
\newline
\begin{figure}[t]
	\begin{center}
		\includegraphics[width=1\linewidth]{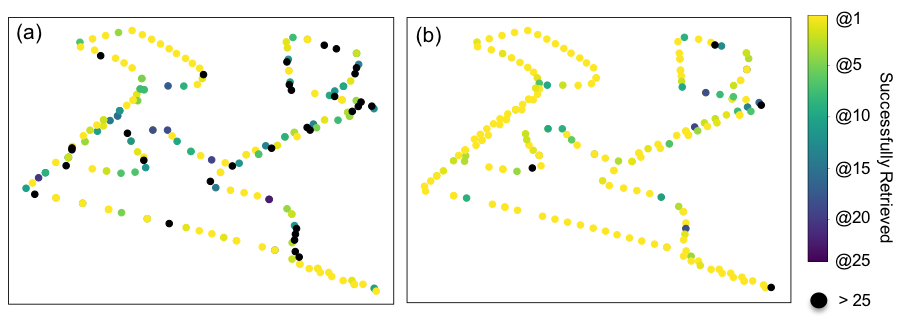}
	\end{center}
	\vspace{-0.4cm}
	\caption{Figure shows the retrieved map of our PointNetVLAD for a randomly selected database-query pair of the unseen B.D. for (a) baseline model and (b) refined model.\vspace{-0.3cm}}
	\label{fig:map}
\end{figure}

\vspace{-0.2cm}
\noindent\textbf{Usability} We further studied the usability of our network for place recognition. Fig. \ref{fig:map} shows heat maps of correctly recognized submaps for a database pair in B.D. before and after network refinement. The chosen database pair is the pair with the lowest initial recall before network refinement. It is shown that our network indeed has the ability to recognize places almost throughout the entire reference map. Inference through our network implemented on Tensorflow\cite{45381} on an NVIDIA GeForce GTX 1080Ti takes $\sim9$ms and retrieval through a submap database takes $O(\log n)$ making this applicable to real-time robotics systems.
\newline

%
%
	\vspace{-0.6cm}
\section{Conclusion}
\vspace{-0.2cm}
We proposed the PointNetVLAD that solves large scale place recognition through point cloud based retrieval. 
We showed that our deep network is permutation invariant to its input. We applied metric learning for our network to learn a mapping from an unordered input 3D point cloud to a discriminative and compact global descriptor for the retrieval task. 
Furthermore, we proposed the ``lazy triplet and quadruplet" loss functions that achieved more discriminative and generalizable global descriptors. Our experimental results on benchmark datasets showed the feasibility and usability of our network to the largely unexplored problem of point cloud based retrieval for place recognition.\newline

\noindent\textbf{Acknowledgement} We sincerely thank Lionel Heng from DSO National Laboratories for his time and effort spent on assisting us in data collection.
	
	{\small
		\bibliographystyle{ieee}
		\bibliography{ref}
	}

	\vspace{+4cm}
	\appendix
	\noindent \section*{Supplementary Materials}
	\section{Benchmark Datasets}
	\begin{figure*}
		\begin{center}
			\includegraphics[width=1\linewidth]{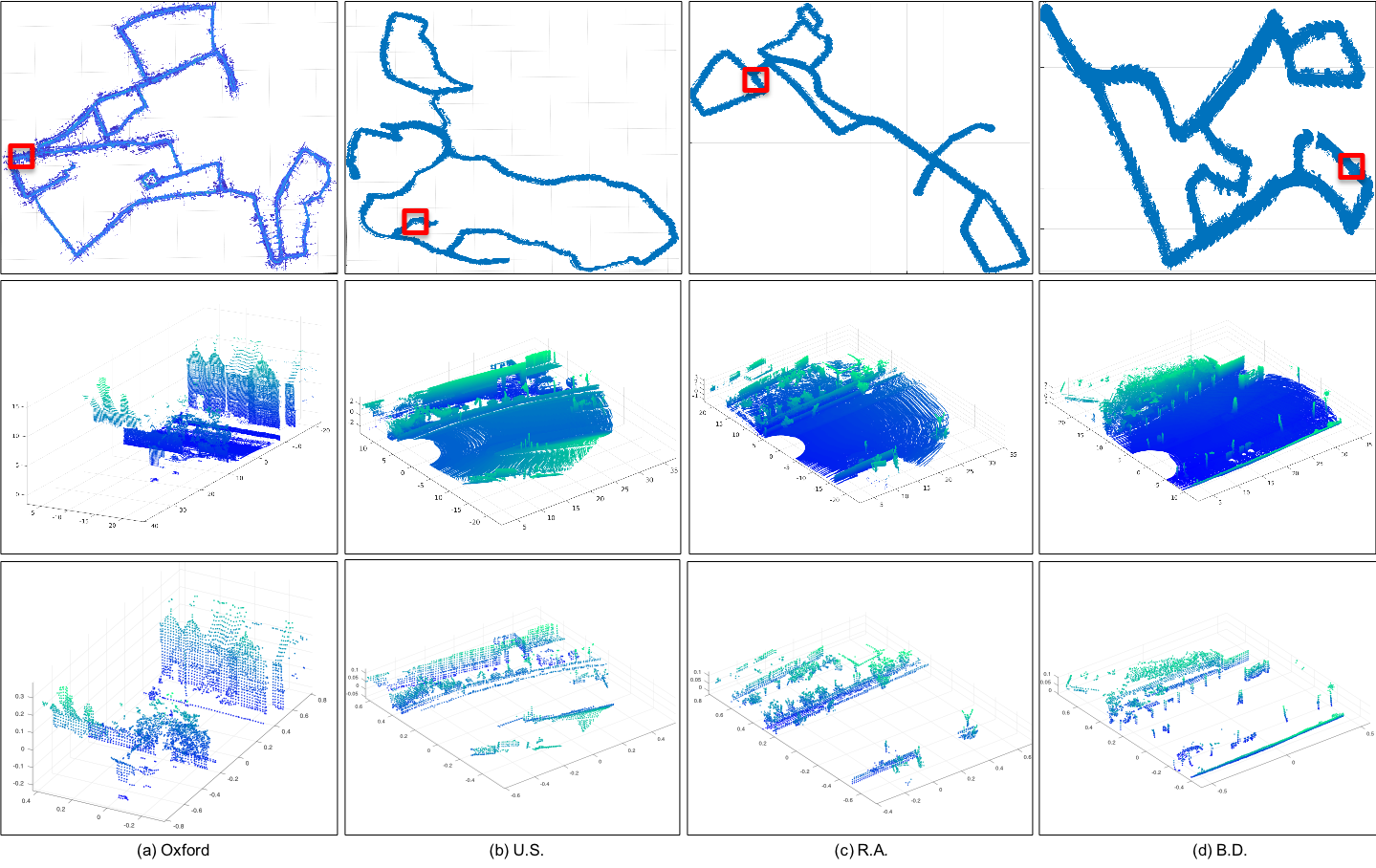}
		\end{center}
		\caption{Top row shows a sample reference map from (a) Oxford, (b) U.S., (c) R.A and (d) B.D.. Middle row shows a sample submap from each of the regions representing the local area marked by the red box on the reference map. Bottom row shows the corresponding preprocessed submaps of the local areas from the middle row. }
		\label{fig:map}
	\end{figure*}
\begin{figure*}
		\begin{center}
			\includegraphics[width=1\linewidth]{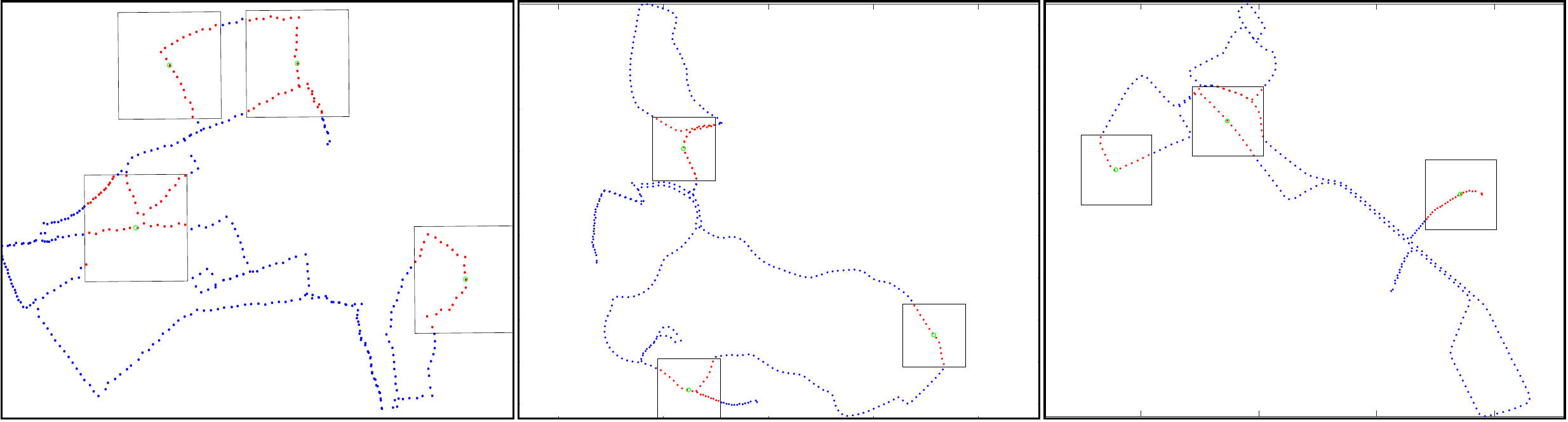}
		\end{center}
		\caption{Data splitting: Blue points represent submaps in the training reference map and red points represent submaps in the testing reference map. The data split was done by randomly selecting regions in the full reference map. }
		\label{fig:disjointmap}
	\end{figure*}

	We provide additional information on the benchmark datasets that are used to train and evaluate the network, which are based on the Oxford RobotCar dataset \cite{RobotCarDatasetIJRR} and three in-house datasets. Figure \ref{fig:map} (top row) shows a sample reference map for each of the four regions. Figure \ref{fig:map} (middle row) shows sample submaps from the different regions. Figure \ref{fig:disjointmap} illustrates data splitting into disjoint reference maps, which was done by randomly selecting 150m $\times$ 150m regions.
	\newline

	\section{Implementation Details}
	We use a batch size of 3 tuples in each training iteration. Each tuple is generated by selecting an anchor point cloud $P_a$ from the set of submaps in the training reference map followed by an on-line random selection of $P_{pos}$ and $\{P_{neg}\}$ for each anchor. Each training tuple contains 18 negative point clouds, i.e. $|\{P_{neg}\}|=18$. Hard negative mining is used for faster convergence by selecting the hardest/closest negatives from 2000 randomly sampled negatives to construct $\{P_{neg}\}$ for each anchor $P_a$ in an iteration. The hard negatives are obtained by selecting the 18 closest submaps from the cached global descriptors $f$ of all submaps in the training reference map, and the cache is updated every 1000 training iterations. We also found network training to be more stable when we take the best/closest of 2 randomly sampled positives to $P_a$ in each iteration.
	
\end{document}